\documentclass[graybox]{support/svmult}

\usepackage{booktabs}
\usepackage{newtxtext}       
\usepackage{helvet}         
\usepackage{courier}        
\usepackage{makeidx}         
\usepackage{multicol}        
\usepackage[bottom]{footmisc}

\usepackage{easy-todo}
\usepackage{float}
\usepackage[square,sort,comma,numbers]{natbib}  

\usepackage{enumerate}
\usepackage[pdftex]{graphicx}
\usepackage[pagebackref=true,breaklinks=true,letterpaper=true,colorlinks,citecolor=black,bookmarks=false]{hyperref}
\usepackage[font={footnotesize}, skip=1pt]{caption}
\usepackage{subcaption}
\usepackage{amsmath,amssymb,amsopn,amstext,amsfonts}
\usepackage{url}
\usepackage{pdfsync}
\usepackage{balance}
\usepackage[capitalise]{cleveref}
\usepackage[table]{xcolor}
\usepackage{algorithm}
\usepackage{algpseudocode}
\usepackage{multirow}
\usepackage{mathrsfs}
\usepackage{isomath}
\usepackage{gensymb}
\usepackage{verbatim}
\usepackage{makecell}

\usepackage{titlesec} 

\titlespacing*{\section}{0pt}{0.4\baselineskip}{0.2\baselineskip}
\titlespacing*{\subsection}{0pt}{0.4\baselineskip}{0.2\baselineskip}
\titlespacing*{\subsubsection}{0pt}{0.4\baselineskip}{0.2\baselineskip}

\usepackage{./support/macro}

\DeclareGraphicsExtensions{.pdf,.png,.jpg}
\graphicspath{{media/}}

\makeindex             

\begin{document}

\title*{Fast and Agile Vision-Based Flight with Teleoperation and Collision Avoidance on a Multirotor}
\titlerunning{Agile Vision-Based Flight and Collision Avoidance}
\author{Alex Spitzer$^*$, Xuning Yang$^*$, John Yao, Aditya Dhawale, Kshitij Goel, Mosam Dabhi, Matt Collins, Curtis Boirum, and Nathan Michael}
\authorrunning{A. Spitzer \textit{et al.}}
\institute{\scriptsize Authors are with the Robotics Institute at Carnegie Mellon University, 5000 Forbes Ave, Pittsburgh PA, 15213. \{aspitzer,xuningy,johnyao,adityand,kgoel1,mdabhi,mcollin1,cboirum,nmichael\}@andrew.cmu.edu\\ $^*$ These authors contributed equally.}
%
%
\maketitle

\vspace{-1em}
\abstract{
We present a multirotor architecture capable of aggressive autonomous flight and collision-free teleoperation in unstructured, GPS-denied environments.
The proposed system enables aggressive and safe autonomous flight around clutter by integrating recent advancements in visual-inertial state estimation and teleoperation.
Our teleoperation framework maps user inputs onto smooth and dynamically feasible motion primitives. Collision-free trajectories are ensured by querying a locally consistent map that is incrementally constructed from forward-facing depth observations.
Our system enables a non-expert operator to safely navigate a multirotor around obstacles at speeds of $10$ m/s.
We achieve autonomous flights at speeds exceeding $12$ m/s and accelerations exceeding $12$ m/s$^2$ in a series of outdoor field experiments that validate our approach.
}

\section{Introduction}

Autonomous aerial vehicles with onboard vision-based sensing and planning have been shown to be capable of performing fast and agile maneuvers.
However, long-horizon planning required to achieve a task has proven to be a challenge, particularly with limited onboard compute.
We propose a fully integrated vision-based multirotor platform that allows minimally trained operators to teleoperate the vehicle at high speeds with agility, while remaining safe around obstacles in unstructured outdoor environments.
At high speeds, the environment around the vehicle changes quickly, and is subject to dynamic obstacles and lighting conditions.
Our multirotor architecture integrates the following to achieve agile and collision-free flight in these scenarios: 1) an extension of motion primitive-based teleoperation \cite{Yang2018} to generate jerk-continuous local trajectories, a crucial component to prevent instability in agile flights, and 2) efficient local mapping for collision avoidance purposes using a KD-tree.

Many prior works in high speed flight exploit the field-of-view (FOV) of stereo cameras for fast collision checking for autonomous flights, including \cite{florence2016, florence2017}, where trajectories are constrained to be inside the FOV of the depth sensor with a max range of $10$ m.
In \cite{matthies2014}, trajectories generated by RRT$^*$ are checked for collisions directly in the disparity space.
\citet{lopez2017} presents aggressive flight on a $1.2$ kg MAV, achieving a velocity of $3$ m/s.
These methods achieve fast collision checking by circumventing the need to construct a local map and checking for collisions in the sensor's FOV.
This however, limits the range of motions the vehicle can perform.
Approaches with local map generation using a laser range finder \cite{chen2016} and a monocular RGB camera \cite{daftry2016} have been shown to achieve maximum velocities of $1.8$ m/s and $1.5$ m/s respectively, but data processing limits the update rate of the local maps.
In our approach, we give a minimally trained operator full control of the vehicle and show that fast and agile flights can be achieved with a human-in-the-loop while maintaining safety.

We perform a series of high speed collision avoidance trials in both indoor and outdoor environments with untrained operators. In our experiments, our hexarotor attains speeds exceeding $12$ m/s and accelerations exceeding $12$ m/s$^2$. We are able to safely avoid obstacles at speeds up to $10$ m/s and accelerations of $8$ m/s$^2$, while retaining a local map.

\section{Technical Approach}

\subsection{Smooth Motion Primitive-Based Teleoperation}
Aggressive multirotor flights demand large angular velocities and large angular accelerations, which are directly related to the jerk and snap of the reference position \cite{mellinger_minimum_2011}.
Thus to avoid incurring large tracking error due to discontinuous trajectories, we extend \textit{forward-arc motion primitives} \cite{Yang2018} to generate trajectories that retain differentiability up to jerk and continuity up to snap.
From the resulting trajectories, we can calculate desired vehicle attitude, angular velocity, and angular acceleration for use as feedforward terms in the controller.

The motion primitives are parameterized as follows.
We define a \textit{local frame} $\Lc$ to be a fixed $z$-axis aligned frame, taken at a snapshot in time.
The motion primitive definition will be provided in the local frame at the time at which an input is issued, and can be freely transformed into a fixed global frame or body frame for control purposes.
An action $\av$ specified by a continuous external input, such as a joystick, generates a single motion primitive.
For a multirotor, we define an action as $\av = \{v_x, v_z, \omega\}$ in the local frame at which the input is issued, where $v_x$ is the $x$-velocity (i.e., the forward velocity of the vehicle), $v_z$ is the $z$-velocity, and $\omega$ is the yaw rate. Let $\xv$ denote a vector containing the position and yaw of the vehicle, i.e. $\xv = [x, \; y, \; z, \; \theta]$. Then, the endpoint velocities are defined by the unicycle model~\cite{Pivtoraiko2009}. The unicycle model dynamics are given by
$
\dot \xv(\av, \tau) = [v_x \cos (\omega \tau) \quad
                           v_x \sin (\omega \tau )\quad
                           v_z  \quad
\omega]\transpose$, where ${\tau \in[0,T]}$. The position endpoints are unconstrained and we enforce all higher order derivatives above velocity to be zero.

A regeneration step $k$ occurs when a new input is received from the joystick, or the previous trajectory $\gammav(\av_{k-1}, T)$ finishes executing.
Alternatively, a fixed regeneration rate can be chosen in order to accommodate changes in the environment for collision avoidance.
Suppose regeneration step $k$ occurs at time $t_k$. Then, a library of dynamically feasible motion primitives is generated in the local frame $\Lc_{t_k}$ specified by the reference state at time $t_k$, i.e. $\xv_{\text{ref}}(t_k) = \gammav(\av_{k-1}, t_k - t_{k-1})$, given a set of discretized actions $\av_k$. Each motion primitive $\gammav$ is a vector of four $8^\text{th}$ order polynomials that specify the trajectory along the position coordinates $x$, $y$, $z$ and yaw coordinate $\theta$. Given an action $\av_k$ at regeneration step $k$, each motion in the motion primitive library is generated in frame $\Lc_{t_k}$ according to
\begin{align} \label{eq:traj}
  \gammav(\av_k, \tau) & = \sum_{i=0}^{8} \cv_i \tau^i \\
  \st & \gammav^{(j)}(\av_k, 0) = \xv_{\text{ref}}^{(j)} (t_k) \;\; \text{for } j = 0, 1, 2, 3, 4 \nonumber \\
  & \dot \gammav(\av_k, T) = \dot \xv (\av_k, T) \nonumber \\
  & \gammav^{(j)}(\av_k, T) = 0 \;\; \text{for } j = 2, 3, 4 \nonumber
\end{align}
where $\{\cdot\}^{(j)}$ specifies the $j^\text{th}$ time derivative.
Note that all constraints are appropriately transformed into $\Lc_{t_k}$.
The duration of the trajectory $T$ and the maximum $x$, $z$, and yaw velocities are specified according to the desired operational range.
We further allow the operator to freely rotate the motion primitive library with respect to $\Lc_{t_k}$'s $z$-axis to allow for sideway slalom motions.

\begin{figure}
   \centering
   \captionsetup[subfigure]{position=b}
   \begin{tabular}[t]{cc}
   \subcaptionbox{\label{fig:cont_traj}}{\includegraphics[width=0.32\linewidth]{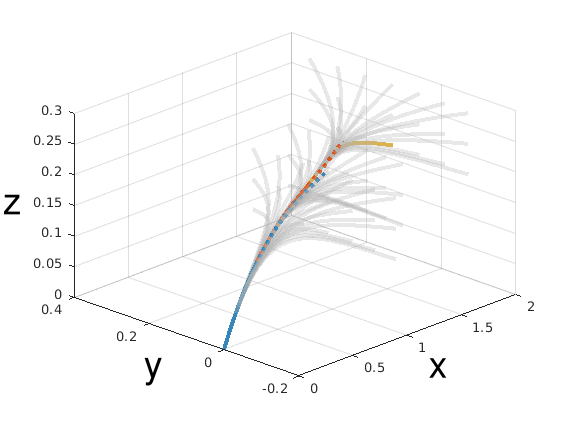}} &
   \subcaptionbox{\label{fig:cont_hod}}{\includegraphics[trim={1.0in, 0, 1in, 0}, clip, width=0.68\linewidth]{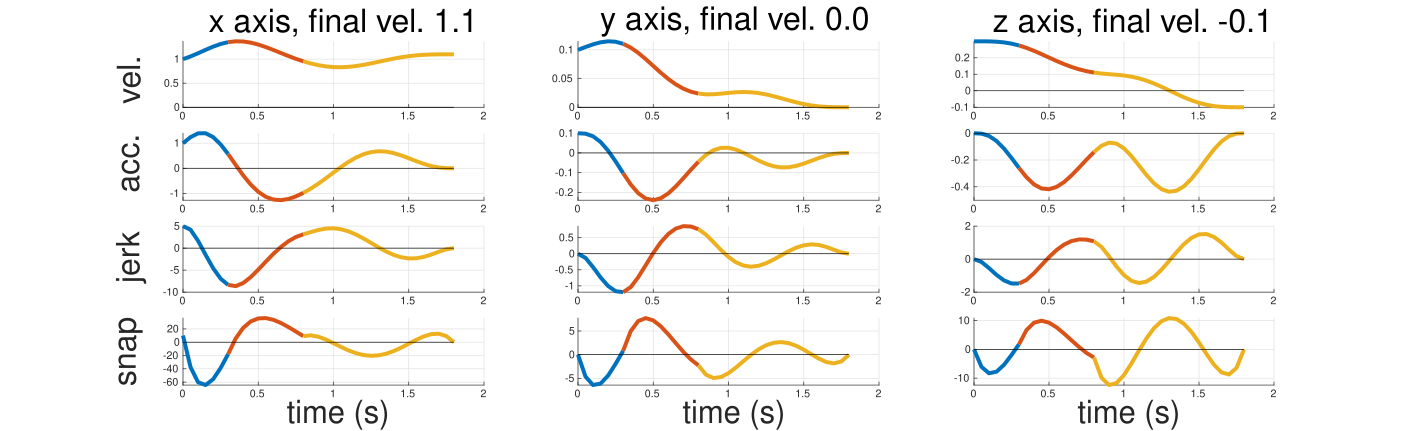}}\\
  \end{tabular}
  \caption{(a) A trajectory composed of 3-segments of motion primitives that switches to a new motion primitive at arbitrary points along the trajectory that have non-zero higher-order-derivative terms. The discarded trajectory is shown in dotted lines. (b) Higher-order time derivatives (velocity, acceleration, jerk, and snap) of the three segment trajectory, showing that the trajectories are differentiable up to jerk and continuous in snap at the switching points. At the end of the trajectory, all higher order derivatives are zero except for the user specified velocity.\label{fig:cont}}

\end{figure}

The result of having snap-continuous trajectories (see Fig.~\ref{fig:cont}) ensures that we have smoothness in error dynamics, thus minimizing instabilities and tracking error.

\setlength{\tabcolsep}{0pt}

\begin{figure}
   \centering
   \captionsetup[subfigure]{position=b}
   \begin{tabular}[t]{cc}
     \subcaptionbox{\label{fig:varying_in_x}}{\includegraphics[width=0.4\linewidth]{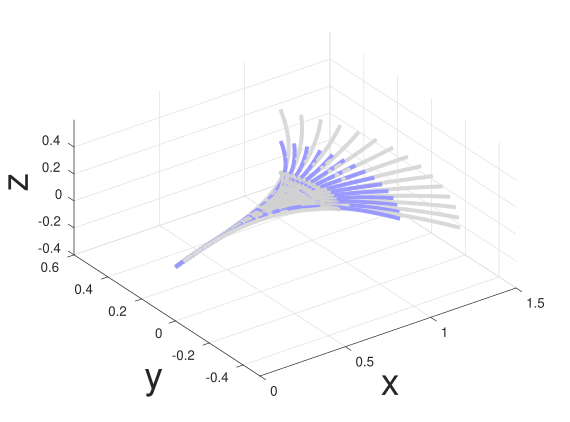}} &
   \subcaptionbox{\label{fig:varying_in_y}}{\includegraphics[width=0.4\linewidth]{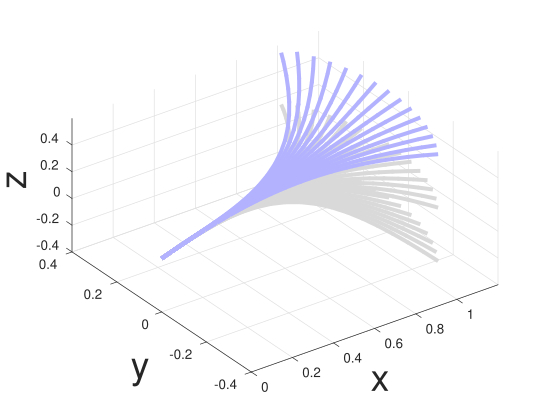}}\\
  \end{tabular}
  \caption{Motion primitive library with (a) variation in linear velocity and with (b) variation in $z$ velocity \label{fig:varying_mp}}
\end{figure}

\subsection{Pruning \& Trajectory Selection.}
At every time step, a family of motion primitives, called the motion primitive library, is created.
The motion primitive library is constructed by discretizing the continuous input along each action dimension, such that each action $\av_i \in A$ is selected from a convex set $A := \{ \av \in [\av_\text{min}, \av_\text{max}]\}$ with size $N_1 \times N_2 \times \dots \times N_n$ where $N_i$ is the dimension of the space of each input. An example of the discretization is shown in Fig.~\ref{fig:varying_mp}.

At every time step, the operator input is mapped to the closest input in the action space, as defined by the Euclidean norm. A priority queue that minimizes input distance from the selected input $\av_{\text{joystick}}$ to each input in the action space $\av_i \in A$ is used to iterate through the action space until a feasible, collision-free trajectory is found.
This results in having the operator input mapped to the feasible motion primitive in the library that is parameterized by the closest discretized action, i.e. $\gammav(\av_i) = \argmin \norm{\av_{\text{joystick}} - \av_i}$.
A trajectory is deemed collision-free if the minimum distance between any point along the trajectory and the surrounding environment is above the sum of the vehicle size and an operator specified collision radius.
Algorithm~\ref{alg:pruning} describes teleoperation with reactive collision avoidance in detail.

The effect of this pruning algorithm is that the vehicle exhibits natural behavior in the presence of obstacles.
If a pillar is in front of the vehicle, then the vehicle chooses a motion primitive some angle away and avoids the obstacle.
If a wall is present, then the vehicle will choose linear velocities that gradually decrease until the vehicle is stopped.

\setlength{\textfloatsep}{4pt}
\begin{algorithm}[htb!]
  \footnotesize

	\caption{Snap-continuous Motion Primitives based Teleoperation and Collision Checking with KD-Tree Local Map}\label{alg:pruning}
	\begin{algorithmic}[1]
    \small
    \State \textbf{Given} KD-Tree local map \textit{L}, collision radius $r$, vehicle radius $r_v$
    \State Receive input $\av_{\text{joystick}}$
    \State Generate the minimum input distance queue according to $d_i = \norm{\av_{\text{joystick}} - \av_i}$
    \While {$\av_i$ is infeasible}
      \State Pop the top action element off of the minimum input distance queue $\av_i$
      \State Generate $\gammav_i = \gammav(\av_i)$ according to \eqref{eq:traj}
      \For {$\tau = 0:T$ discretized at some $\triangle t$}
        \State Query \textit{L} for the closest point $p$ to $\gammav_i(\av, \tau)$
        \If{$\|p - \gammav_i(\av, \tau) \| \geq r + r_v $}
            \State Set $\av_i$ to feasible
            \State Set $\gammav = \gammav_i$
          \EndIf
			\EndFor
    \EndWhile
	\end{algorithmic}
\end{algorithm}

\subsection{Local Map Representation using KD-Trees}

We present a local mapping framework that generates a spatially consistent local map 
of the robot surroundings represented as voxel grids. The local map is generated by 
retaining only the depth sensor measurements obtained at poses that lie in the
vicinity of the vehicle's current pose. This enables trajectories to span in 
the space observed by all of the retained sensor measurements. Sequential sensor 
measurements obtained using a stereo imaging sensor often contain redundant information 
about the surroundings of the robot. The novelty of information in an incoming 
sensor measurement at the resolution of the voxel grids is dictated by the rotation and 
translation displacement of the robot between the current frame and the previous frames. 
To enforce spatial locality, we dynamically select anchor frames and transform 
subsequent sensor measurements that contain novel information about the surroundings 
in to the anchor frames.
In order to efficiently incorporate only novel information in the local map,
we classify each incoming sensor measurement into one of the following categories: a 
KeyFrame (\textit{KF}), a SubFrame (\textit{SF}), or a BufferFrame (\textit{BF}) 
(see Table~\ref{table:frame_classification}).
The local map (\textit{L}) is updated in a locally consistent coordinate frame 
according to the type of sensor frame 
($\textit{F} \in \{\text{\textit{KF}, \textit{SF}, \textit{BF}}\}$). 

A sensor measurement that is more than $\alpha_k$ meters away from the current \textit{KF} is classified as a \textit{KF}. Sensor measurements that are not new \textit{KF}s, but are either more than $\alpha_s$ meters in position or $\beta_s$ degrees in heading away from the previous \textit{SF}, are classified as \textit{SFs}.
Sensor measurements that are neither \textit{KF}s nor \textit{SF}s are classified as \textit{BF}s, which do not contain sufficient novel information about the surroundings, but are registered to \textit{L} to account for dynamic 
changes in the scene. \textit{BF}s are in \textit{L} only for the iterations in which they are observed.

\begin{table}[h]
  \centering
  \caption{Frame Classification}
    \begin{tabular}{c|l|l}
      \toprule
    {\textbf{ Frame Type }}& {\textbf{ Definition}} &{\textbf{ Initialization Condition for Frame} $\bm{F}$} \\
      \hline
      \textit{KF}& Sensor frames to which subsequent & $ T \left(\mathit{KF_{i-1}}, \mathit{F}\right) > \alpha_k $\\
      (Anchor)& \textit{SFs} and \textit{BFs} are registered & \\
      \hline
      \textit{SF}& Sensor frames that provide novel information & $T\left(\mathit{SF_{i-1}}, \mathit{F} \right) > \alpha_s|| $ \\
      & about the vehicle surroundings& $H\left(\mathit{SF_{i-1}}, \mathit{F}\right) > \beta_s$\\
      \hline
      & Sensor frames that are registered for one time& \\
      \textit{BF}& step to accommodate the dynamic changes & $\if \mathit{F} \neq \left(\mathit{KF}, \mathit{SF}\right)$\\
      & in the surroundings when the vehicle has & \\
      &  not been displaced by a sufficient distance & \\
      \bottomrule
      \multicolumn{3}{l}{$T\left(a, b\right)$ is the translational distance between $a$ and $b$; $H\left(a, b\right)$ is the heading rotational distance}\\
      \multicolumn{3}{l}{between $a$ and $b$.}
    \end{tabular}\label{table:frame_classification}
\end{table}

A new local map \textit{L} is spawned every time a new \textit{KF} is detected.
\textit{L} consists of all the \textit{SF}s registered to the current \textit{KF}, 
the current \textit{BF} and all the \textit{SF}s registered to the previous \textit{KF}, 
resulting in a voxel grid map representing the occupied space centered around the vehicle.
This allows us to only update the KD-Tree that contains the voxel centers of the 
\textit{KF} to which the incoming sensor scan is registered.
Because a high fidelity map is not essential for collision avoidance, each sensor 
measurement is downsampled into an occupancy grid with a fixed voxel size before 
registering it to \textit{L}.
Voxel centers in the map are then arranged in a KD-Tree to minimize the time complexity of 
nearest neighbor queries.

The algorithmic efficiency and independence of the definition of the local map from the 
sensor model enables our approach to incorporate measurements from 
multiple depth sensors with widely separate FOVs.

\subsection{State Estimation}

We use VINS-Mono \cite{Qin18_TRO}, a tightly-coupled visual-inertial odometry framework that has been shown to perform favorably when compared to other state of the art open source state estimation algorithms \cite{Delmerico18_ICRA}.
VINS-Mono jointly optimizes vehicle motion, feature locations, camera-IMU extrinsics, and IMU biases over a sliding window of monocular images and preintegrated IMU measurements.
Because our local planning strategy does not require a globally consistent map, we disable the loop closure functionality of VINS-Mono to reduce its computational footprint.
We run an auxiliary state estimator during takeoff in order to provide smooth state estimates when the vehicle has not yet experienced sufficient motion excitation for VINS-Mono to initialize.
The auxiliary state estimator is an unscented Kalman filter that fuses downward rangefinder altitude observations, downward optical flow, and IMU measurements to estimate vehicle velocity, position, and attitude.

\section{Implementation} \label{sec:implementation}

\begin{figure}
   \centering
   \captionsetup[subfigure]{position=b}
   \begin{tabular}[t]{ccc}
     \subcaptionbox{\label{fig:vehicle}}{\includegraphics[height=0.27\linewidth]{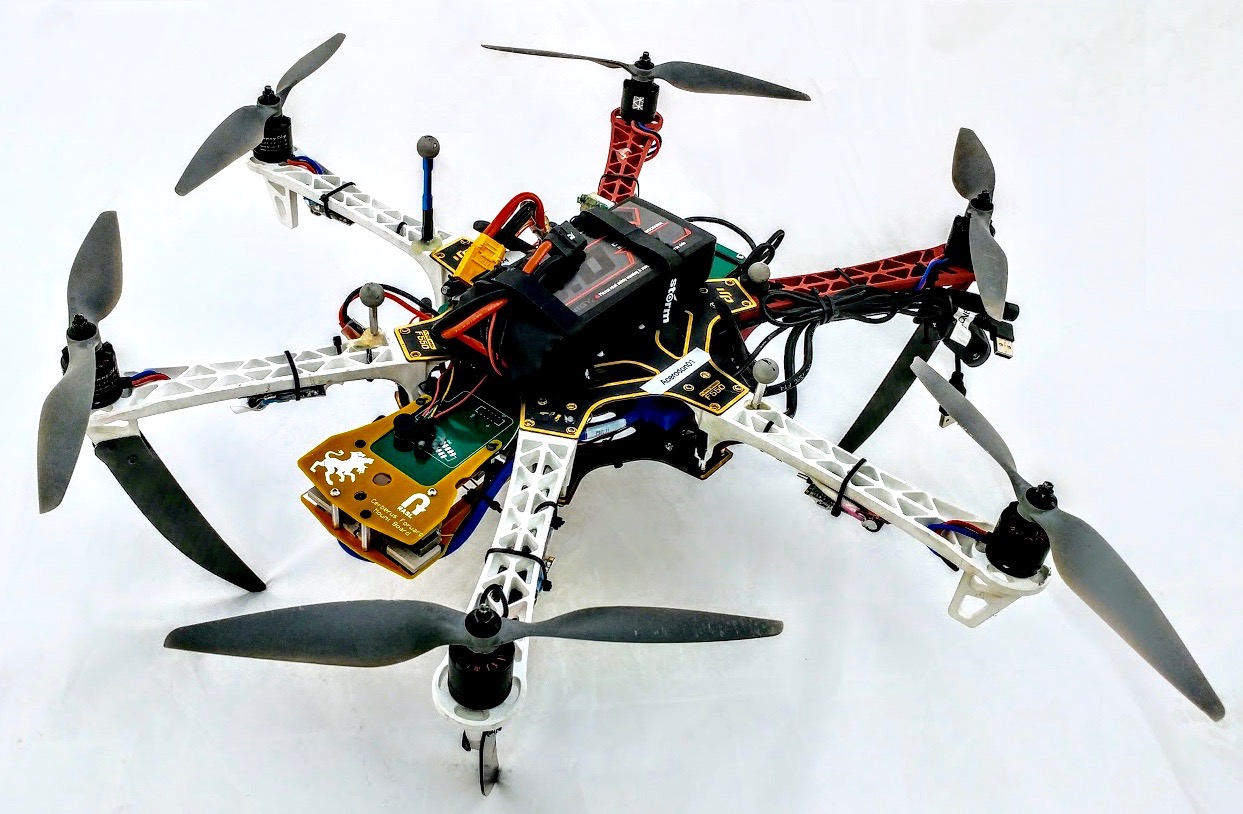}} &
       \subcaptionbox{\label{fig:outdoor_overlay}}{\includegraphics[trim={3em, 0em, 3em, 0em}, clip, height=0.27\linewidth]{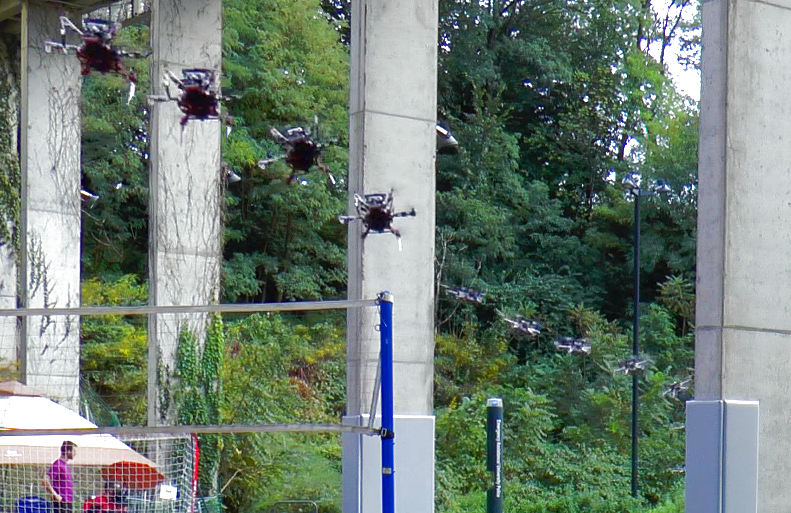}} &
   \subcaptionbox{\label{fig:indoor_overlay}}{\includegraphics[trim={0, 10em, 0, 0em}, clip, height=0.27\linewidth]{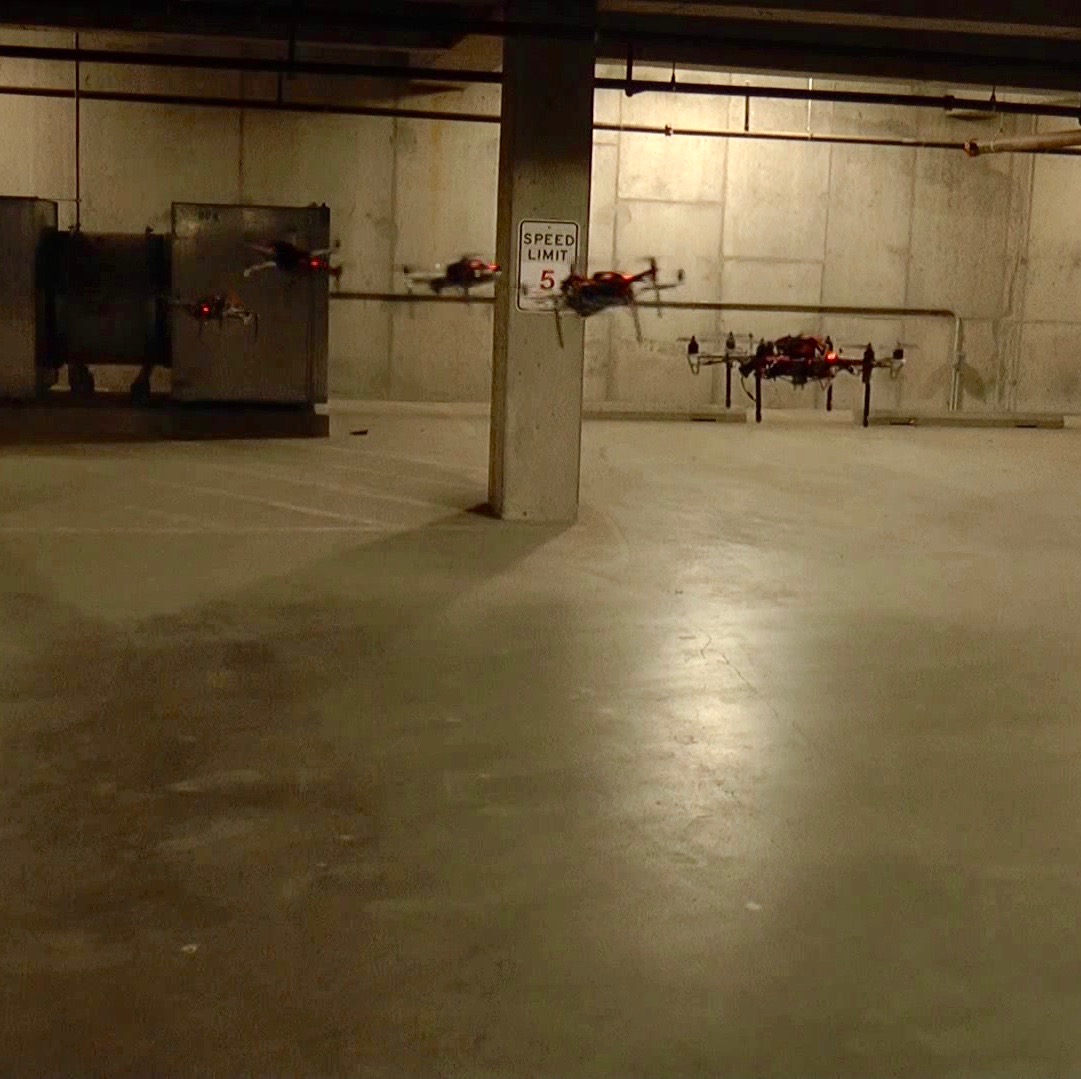}}
  \end{tabular}
  \caption{(a) The hexarotor vehicle used for aggressive flights in (b) outdoor environments and (c) a dimly lit garage. The overlays of the vehicle positions over time depict the trajectories the vehicle took to avoid pillars in its way. \label{fig:overlay}}
\end{figure}

\textit{Hardware.} We experimentally evaluate our proposed approach on a 3.8 kg hexarotor that fits within a 20~cm~$\times$~60~cm~$\times$~80~cm volume (Fig.~\ref{fig:vehicle}).
The hexarotor has an average flight time of 7~min and a power to weight ratio of 3.
Two Intel RealSense D435 depth cameras are used for mapping: one facing forwards and one facing upwards at a 45 degree angle, which aids obstacle avoidance while accelerating forwards.
A downward facing Matrix Vision mvBlueFOX-MLC200w is used as the RGB camera input for VINS-Mono and the NVIDIA Tegra TX2 is used for computation.
The hexarotor uses a cascaded PD control architecture as in \cite{mellinger_minimum_2011} with jerk and snap references used to compute feedforward angular velocities and accelerations.

\textit{Teleoperation.} Motion primitives are generated with an angular velocity bound of $2$ rad/s. There are $25$ discretizations for the $v_x$ action, 11 for $\omega$, and 5 for $v_z$ for a total of $1375$ motion primitives generated per trajectory iteration. Trajectories are generated at $25$ Hz, and the local map is updated at $30$ Hz.

We follow \cite{liu_high_2016} and choose the maximum velocity of the motion primitives to be such that the vehicle can always safely stop given a known constant maximum acceleration, known sensor range and known sensor and mapping rates. Since our vehicle has a power to weight ratio of 3, we assume a conservative maximum acceleration at $6$ m/s$^2$, and assume a worst case sensor and mapping rate of $10$ Hz. With a sensor range of $10$ meters, this allows for a maximum motion primitive velocity of $10.37$ m/s.

\section{Experiments and Results}

\subsection{Experiments in Aggressive High Speed Flights}

We conduct seven aggressive flight experiments in order to test our framework, including four outdoor experiments over an area of $40\text{ m}\times20 \text{ m}$ with obstacles over grass (Fig.~\ref{fig:outdoor_overlay}), and two indoor experiments in a dimly lit garage (Fig.~\ref{fig:indoor_overlay}). The experiments are described in Table~\ref{tab:teleop_params}. The local map used for all experiments uses $\alpha_k=2\text{ m}$, $\alpha_s=0.2\text{ m}$, $\beta_s=0.1\text{ rad}$ and a voxel size of $0.2\text{ m}$.

\begin{table}
  \centering
    \caption{Experiment descriptions and parameters.}
    \begin{tabular}{l|l|r|r|r|r}
      \toprule
      \textbf{Experiment} & \textbf{Description} & $\Tv$ (s)& \makecell[l]{$\vv_x^{\text{max}}$ (m/s)} & \makecell[l]{$\omega^{\text{max}}$ (rad/s)} & $r$ (m)\\
      \midrule
      \textbf{Outdoor-1}& \makecell[l]{High speed teleoperation outside} & $2.2$ & $12.0$ & $2.0$ & N/A\\
      \hline
      \textbf{Outdoor-2,3}& \makecell[l]{Aggressive teleoperation with\\collision avoidance outside} & $2.0$ & $5.0$ & $1.0$ & $0.8$ \\
      \hline
      \textbf{Indoor-1,2}&  \makecell[l]{Aggressive teleoperation with \\collision avoidance in a dimly lit garage}& $1.5$ & $3.0$ & $1.5$ & $0.4$ \\
      \hline
      \textbf{Outdoor-4}& \makecell[l]{High speed, aggressive teleoperation\\with collision avoidance outside} & $1.3^*$ & $7.0$ & $2.0$ & $0.5$ \\
      \hline
      \textbf{Outdoor-5}& \makecell[l]{High speed, aggressive teleoperation\\with collision avoidance outside} & $1.5^*$ & $10.0$ & $2.0$ & $0.2$ \\
      \bottomrule
      \multicolumn{6}{l}{
    $\Tv$ denotes the duration of the motion primitive, $\vv_x^{\text{max}}$ is the maximum desired speed,}\\
      \multicolumn{6}{l}{$\omega^{\text{max}}$ is the maximum yaw rate, and $r$ is the collision radius.}\\
      \multicolumn{6}{l}{$^*$Motion primitive duration increased adaptively as a linear function of the desired velocity change}
    \end{tabular}\label{tab:teleop_params}
\end{table}

\begin{figure}
\centering
\includegraphics[width=5.3cm]{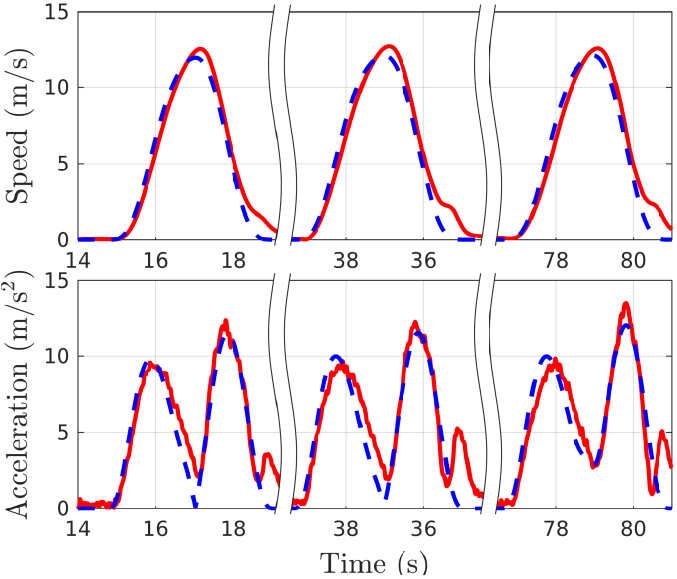}
    \caption{Desired (dashed blue) vs. estimated (solid red) speed and acceleration achieved during high speed flight in experiment \textbf{Outdoor-1}.
A maximum acceleration of 13.5 m/s$^2$ was attained.}
    \label{fig:hs1}
\end{figure}

\emph{High Speed Flight.} In \textbf{Outdoor-1}, we execute straight line trajectories that hit a maximum desired speed of $12$ m/s using the teleoperation system. Figure~\ref{fig:hs1} shows the speed and acceleration attained during three runs. The vehicle achieves a maximum acceleration of 13.5 m/s$^2$.

\begin{figure}
   \centering
  \includegraphics[width=10.5cm]{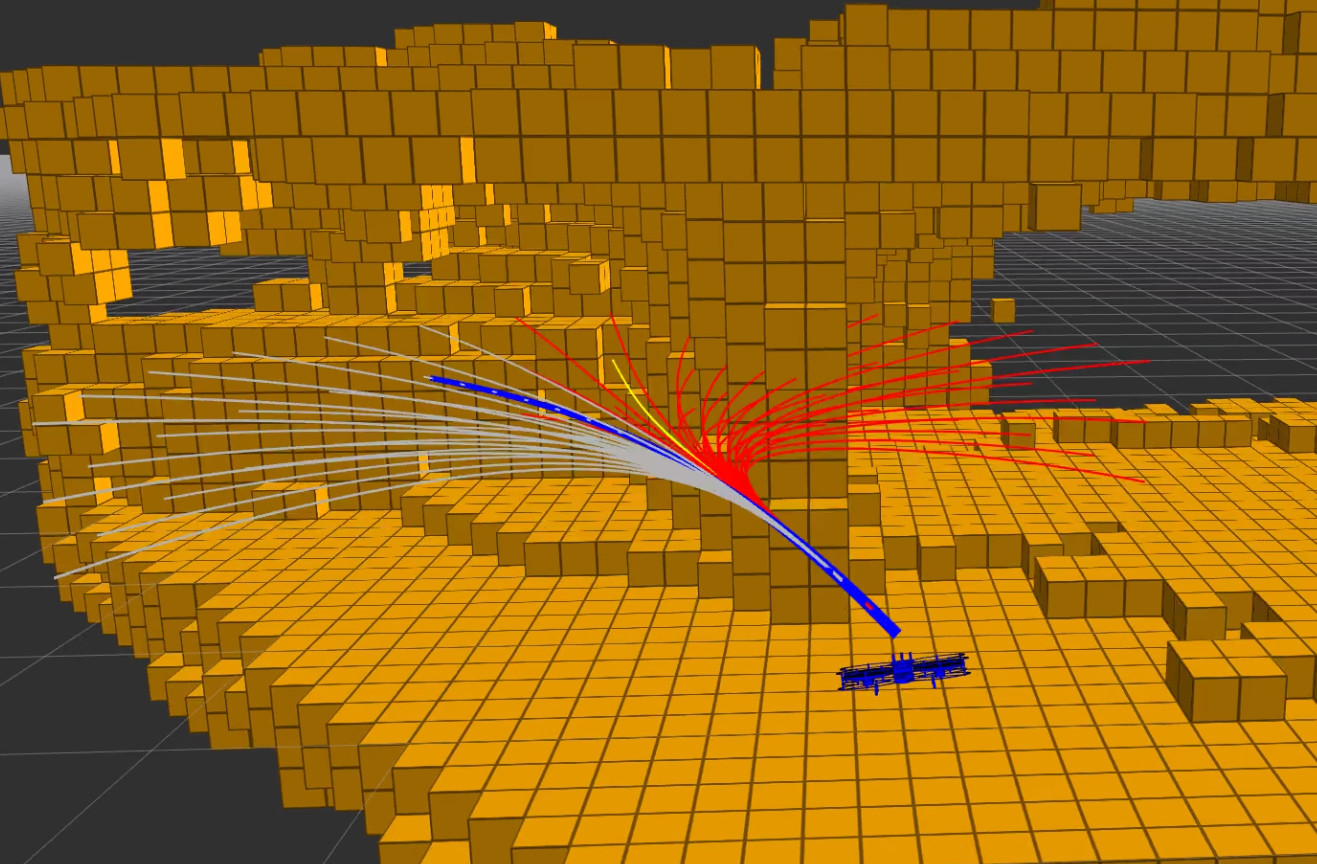}
  \caption{A snapshot of the map and motion primitive library during an indoor experiment when the user selected trajectory (yellow) is not chosen to avoid a collision. \label{fig:map_view}}
\end{figure}

\begin{figure}[H]
   \centering
   \captionsetup[subfigure]{position=b}
   \begin{tabular}[t]{cc}
   \subcaptionbox{\label{fig:O2}}{\includegraphics[width=0.46\linewidth]{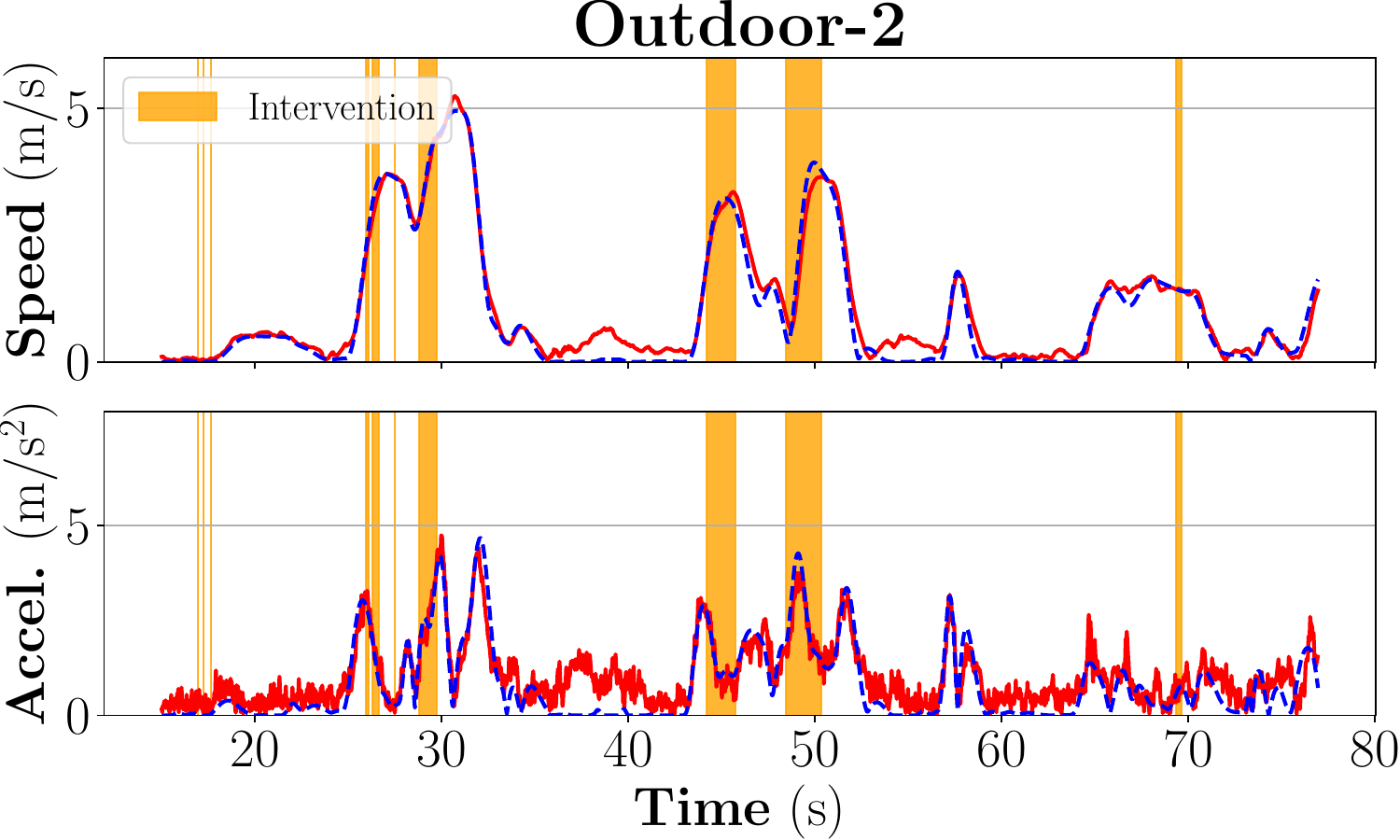}} &
   \subcaptionbox{\label{fig:O3}}{\includegraphics[width=0.46\linewidth]{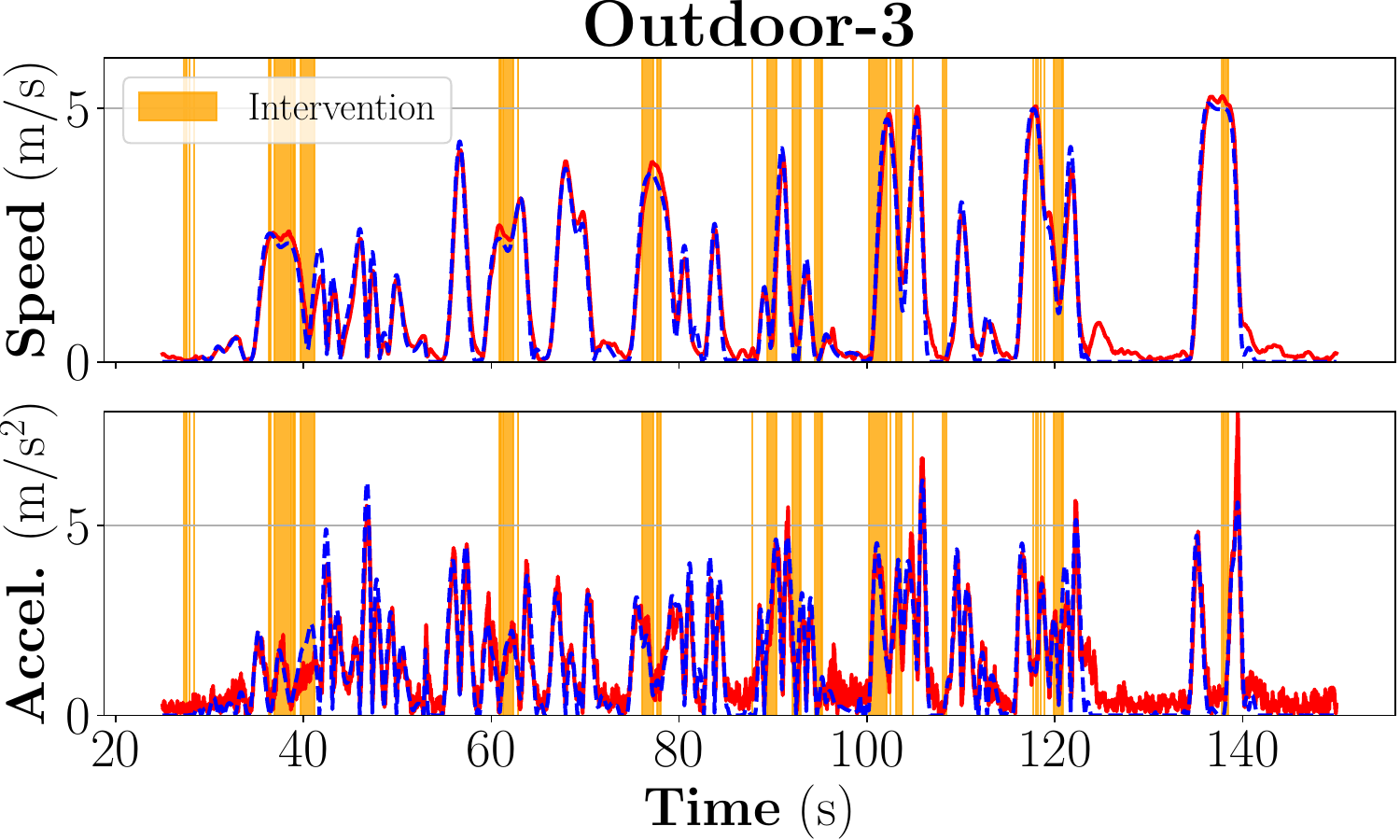}}\\
   \subcaptionbox{\label{fig:I1}}{\includegraphics[width=0.46\linewidth]{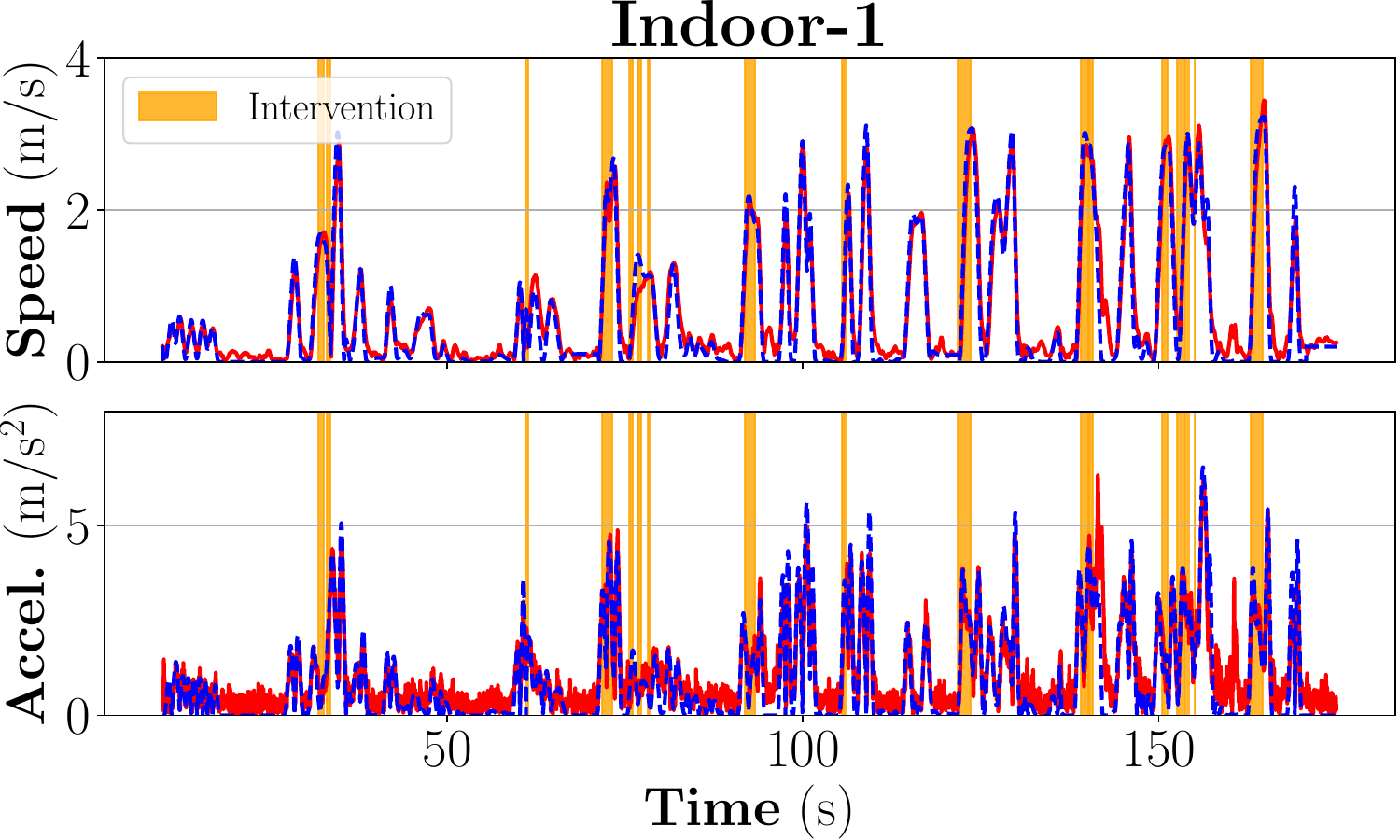}} &
   \subcaptionbox{\label{fig:I2}}{\includegraphics[width=0.46\linewidth]{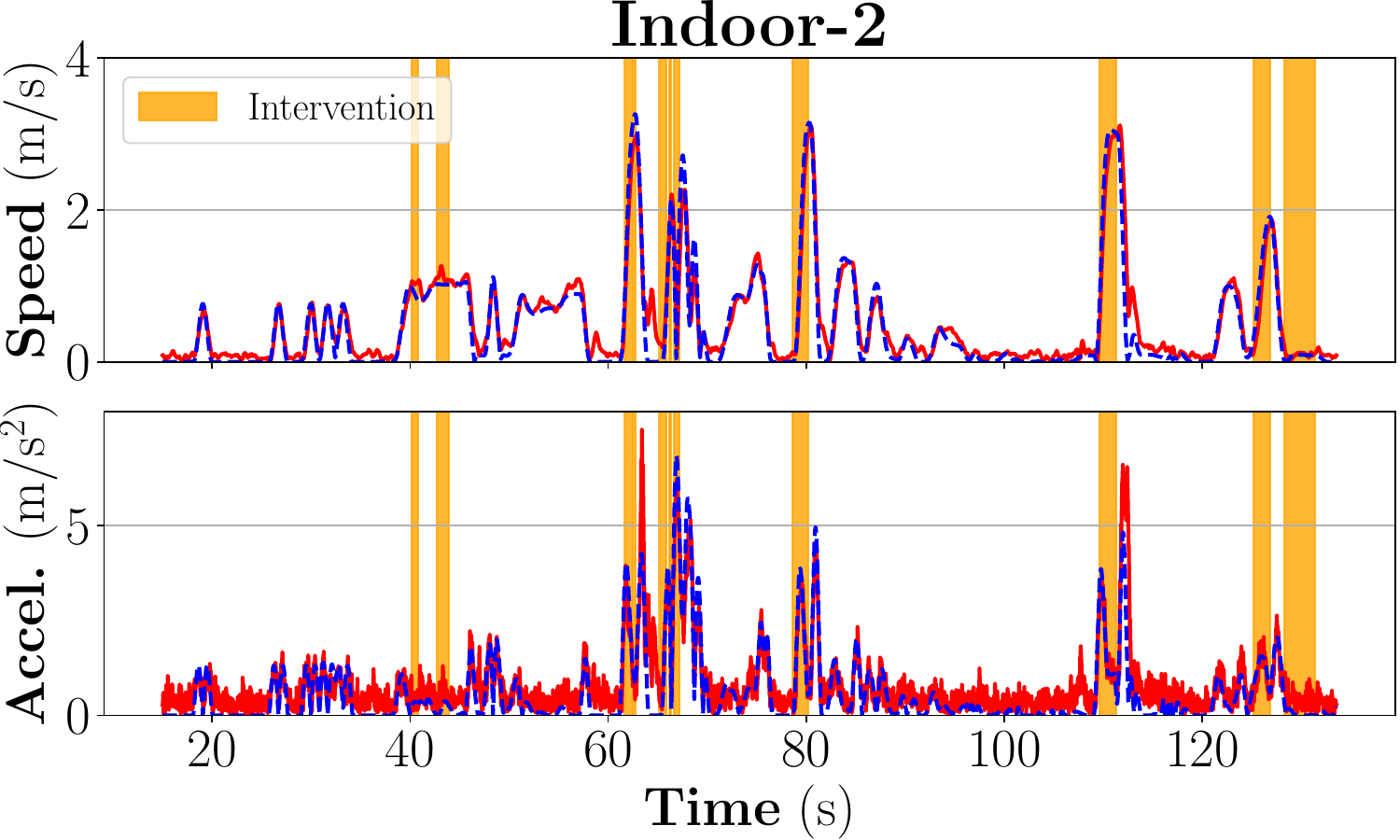}}\\
   \subcaptionbox{\label{fig:I1}}{\includegraphics[width=0.46\linewidth]{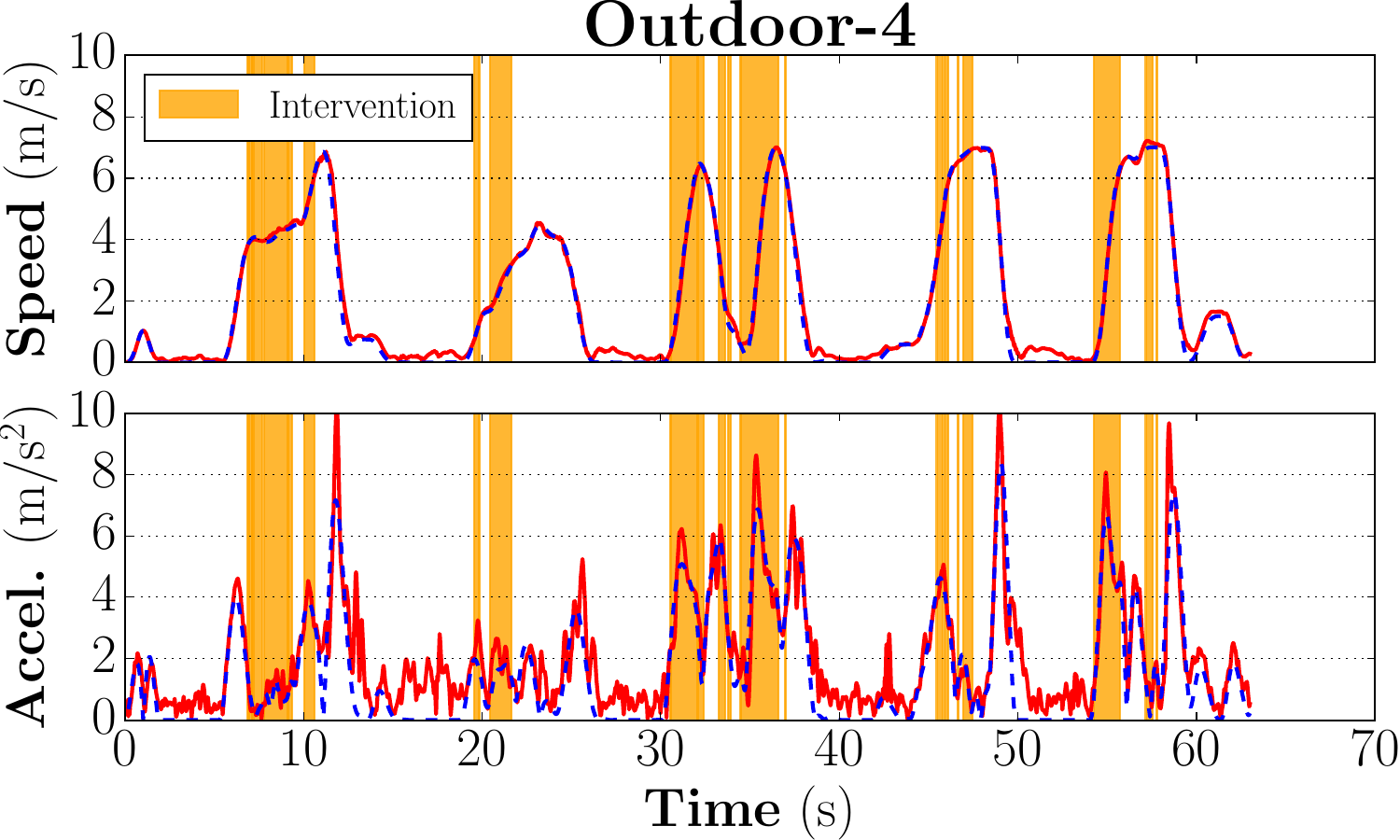}} &
   \subcaptionbox{\label{fig:I2}}{\includegraphics[width=0.46\linewidth]{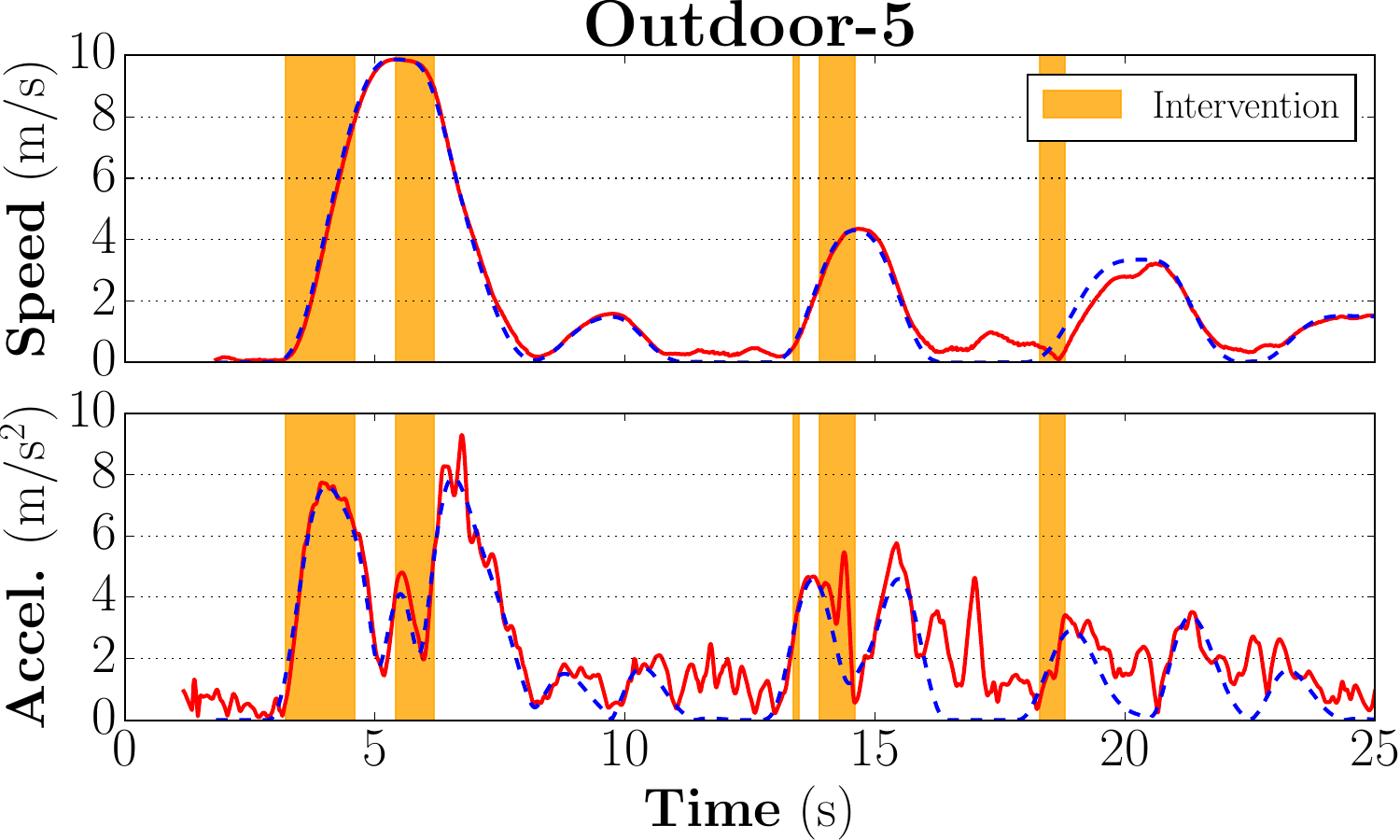}}\\
  \end{tabular}
  \caption{Desired (dashed blue) vs. estimated (solid red) speed and acceleration achieved during our six collision avoidance experiments. During orange regions, the operator's trajectory was not selected in order to keep a minimum distance to surrounding obstacles. The system accurately tracks trajectory references and avoids obstacles while reaching speeds of up to $10\text{ m/s}$, and accelerations up to $8\text{ m/s}^2$. \label{fig:results}}
\end{figure}

\emph{High Speed and Aggressive Flight with Collision Avoidance.}
Experiments \textbf{Outdoor-2-5}, and \textbf{Indoor-1,2} stress test our collision avoidance algorithm at high speeds while maintaining aggressiveness in the commanded trajectories.
In both environments, the operator repeatedly tries to fly the vehicle at the maximum speed, as indicated in Table~\ref{tab:teleop_params}, into an obstacle. Figure~\ref{fig:results} shows the speeds and accelerations attained during the six experiments. The vehicle successfully reaches speeds of $10$ m/s and accelerations of $8$ m/s$^2$ in the outdoor environment and speeds of $3$ m/s and accelerations of $5$ m/s$^2$ in the indoor environment. Figure~\ref{fig:results} also shows regions where the operator's selected trajectory would bring the vehicle closer than $r$ meters to an obstacle. In all such cases, the operator's trajectory is pruned and a collision-free trajectory is selected. Figure~\ref{fig:map_view} shows an example instance of motion primitive pruning to avoid a collision, along with the local map generated by the vehicle.

\emph{Computational Performance Analysis.}
Table~\ref{table:traj_timing} shows that our trajectory generation and pruning time is faster than the user input rate (10 Hz) and that our map generation time is faster than the sensor input rate (30 Hz), enabling real time operation.
Furthermore, Table~\ref{table:overall_cpu} indicates that the in-flight computational footprint of the proposed system is less than 75\% of the available onboard capacity.

\begin{table}[h]
  \centering
  \caption{Execution time (ms) and std. dev. per iteration for safe teleoperation during some of our experiments}
  \resizebox{\textwidth}{!}{%
    \begin{tabular}{c|c|c|c}
      \toprule
      \textbf{Experiment}&{\textbf{Trajectory Generation}}&{\textbf{Trajectory Pruning}}&{\textbf{Local Map Generation} }\\
      \midrule
      \textbf{Outdoor-2}& $0.47 \pm 0.072$ & $29.37 \pm 24.03$ & $5.72 \pm 5.54 $\\
      \textbf{Outdoor-3}& $0.48 \pm 0.071$ & $23.65 \pm 15.42$ & $5.84 \pm 7.95 $\\
      \textbf{Indoor-1}& $0.50 \pm 0.125$ & $8.34  \pm 5.19 $ & $11.67\pm 14.83$\\
      \textbf{Indoor-2}& $0.51 \pm 0.219$ & $15.73 \pm 7.23 $ & $12.71\pm 16.78$\\
      \bottomrule
    \end{tabular}}\label{table:traj_timing}
\end{table}

\begin{table}[h]
  \centering
  \caption{CPU usage (\%, out of a total available 600\%) and std. dev. on a 6 core NVIDIA TX2}
    \begin{tabular}{c|c|c|c}
      \toprule
      \textbf{Experiment}&{\textbf{Safe Teleoperation}}&{\textbf{VINS Mono}}& {\textbf{Total}} \\
      \midrule
      \textbf{Outdoor-2}& $32.92 \pm 17.59$&$67.45 \pm 10.87$ & $395.72 \pm 43.53$ \\
      \textbf{Outdoor-3}& $26.70 \pm 19.98$ & $56.19 \pm 19.47$ & $359.89 \pm 60.87$ \\
      \textbf{Indoor-1}& $57.38 \pm 21.25$ & $64.21 \pm 3.42$  & $450.40 \pm 43.14$ \\
      \textbf{Indoor-2}& $58.76 \pm 24.09$ & $63.82 \pm 6.41$  & $456.47 \pm 36.83$ \\
      \bottomrule
    \end{tabular}\label{table:overall_cpu}
\end{table}

\section{Conclusion and Future Work}

In this work, we present a system architecture designed for collision-free, agile autonomous flight through cluttered environments.
Our hexarotor vehicle is capable of flying at speeds exceeding $12$ m/s and with accelerations exceeding $12$ m/s$^2$, as shown in \textbf{Outdoor-1}.
The teleoperation and collision avoidance system presented has been shown to consistently avoid obstacles while traveling at speeds up to $10$ m/s in an outdoor environment (\textbf{Outdoor-5}) and  in a dimly lit garage (\textbf{Indoor-1}, \textbf{Indoor-2}).

Although we've attained relatively high levels of performance, there are limitations to the architecture.
Currently, motion primitive durations are chosen by the operator.
If the motion primitive duration is set too low, the vehicle will not be able to meet the desired acceleration and performance will suffer.
Consequently, it is difficult to attain high speeds while remaining responsive at low speeds. A natural extension of the approach is to choose motion primitive durations in a principled manner to ensure that generated references are dynamically feasible.
We also believe that chaining together motion primitives to generate more complex reference paths is an exciting avenue for future work. This would enable the vehicle to execute more complicated maneuvers, particularly in cluttered environments.

\bibliographystyle{./support/spmpscinat}
{\scriptsize
\bibliography{refs}
}

\end{document}